\documentclass[10pt,twocolumn,letterpaper]{article}
\pdfoutput=1
\usepackage{scrextend}
\usepackage{cvpr}
\usepackage{times}
\usepackage{epsfig}
\usepackage{graphicx}
\usepackage{subcaption}
\usepackage{amsmath}
\usepackage{amssymb}
\usepackage{array}

\usepackage[breaklinks=true,bookmarks=false]{hyperref}

\cvprfinalcopy 

\def\cvprPaperID{****} 

\ifcvprfinal\fi
\begin{document}

\title{Disentangling Space and Time in Video with Hierarchical Variational Auto-encoders}

\author{Will Grathwohl\\
Lawrence Livermore National Laboratory\\
7000 East Ave, Livermore, CA 94550\\
{\tt\small grathwohl1@llnl.gov}
\and
Aaron Wilson\\
Larwence Livermore National Laboratory\\
7000 East Ave, Livermore, CA 94550\\
{\tt\small wilson256@llnl.gov}
}

\maketitle

\begin{abstract}   
   There are many forms of feature information present in video data. Principle among them are object identity information which is largely static across multiple video frames, and object pose and style information which continuously transforms from frame to frame. Most existing models confound these two types of representation by mapping them to a shared feature space. 
   In this paper we propose a probabilistic approach for learning separable representations of object identity and pose information using unsupervised video data. Our approach leverages a deep generative model with a factored prior distribution that encodes properties of temporal invariances in the hidden feature set. Learning is achieved via variational inference. We present results of learning identity and pose information on a dataset of moving characters as well as a dataset of rotating 3D objects. Our experimental results demonstrate our model's success in factoring its representation, and demonstrate that the model achieves improved performance in transfer learning tasks. 
\end{abstract}

\section{Introduction}

%

Deep neural networks have led to many recent breakthroughs in learning representations of complicated, high dimensional input such as images \cite{krizhevsky2012imagenet}, video \cite{tran2015learning}, text \cite{mikolov2013distributed}, and audio \cite{oord2016wavenet}. However, deep models learn complex feature representations that are not easily understood. While the transferability of deep neural network representations has been studied \cite{yosinski2014transferable}, the underlying factors that affect transferability are are not well known. In most interesting application domains, it is very challenging to obtain enough labeled data to fully train a deep model. Thus, it is of great importance to design models with feature transferability in mind. \cite{higgins2016early} demonstrated that features which decompose into semantic categories can be more successfully transferred to novel tasks since they attend to the high level factors of variation in the input data, instead of just solving the training task. In this paper we investigate a probabilistic approach for learning semantically meaningful image features by exploiting the temporal properties of video data. 



In video (and many other types of sequential data), semantic information such as the identity of a tracked face, or the category of a moving object persists over very long frame sequences. In contrast, the configurations of objects and their locations typically change smoothly and vary frame to frame. We propose an unsupervised approach that learns to separate these distinct semantic categories into separate factors in its representation.

To learn to separate these factors we propose a model of video frame features, inspired by slow feature analysis \cite{wiskott2002slow} that factors the latent representation into two distinct parts; a temporal component that varies smoothly in time and a static component that remains near constant throughout a video sequence. Our approach to modeling these factors is Bayesian. In contrast with most work on slow feature analysis we propose a generative model that encodes the separability assumptions into a prior distribution on the latent state and define a generative distribution for frame contents given the latent representation. Most closely related to our work is \cite{turner2007maximum} which developed a probabilistic interpretation of slow feature analysis. That work proposed a generative framework for learning so called slow features from sequential data. Their model could learn slowly varying features corresponding to the static features learned by our model. Though extensions were proposed in \cite{turner2007maximum} to separately represent static and changing video contents, no efficient means of inferring these decomposed representations were known. We extend their work in two ways; we introduce a prior distribution decomposing the latent state and we propose an efficient method of inferring deep separable representations. 

Our work can be viewed in a second way as an extension of recent contributions that learn meaningful features using variational auto-encoders. Whereas the work on slow feature analysis attempts to encode the concept of invariance into the model, recent work on variational auto-encoders has focused on learning independent features. For example, \cite{higgins2016early} demonstrated that by enforcing independence of features in the latent state it is possible to learn representations that generalize well to new tasks.
Our model combines both concepts into a single probabilistic model. Our empirical results demonstrate that this model is able to learn better feature representations by comparison to models that enforce either invariance or independence alone.  

In this paper we present a hierarchical probabilistic model that learns to decompose the static and temporally varying semantic information in video. We define a prior on the learned frame embeddings which is responsible for the representation's semantic factoring. We describe this prior distribution, explain how inference is performed, and present all derivations needed to train a model that obeys this distribution. We demonstrate that our proposed model successfully learns separable features with human-interpretable meanings. Moreover, we demonstrate, in simple illustrative prediction tasks, the benefit to transfer learning of having learned to disentangle static and temporally varying semantic information. 

    

\section{Preliminaries}
    Our model draws heavily from variational auto-encoders and slow feature analysis. Our approach is inspired by previous work on learning factored and human-interpretable representations. We outline these concepts below.
    
    



\subsection{Slow Feature Analysis}
    Slow Feature Analysis is a technique for learning representations of time-series data motivated by the intuition that high-level semantic information is likely to change at a much slower rate than raw input data. Given an $F$-dimensional time-series $[x_i, ..., x_N]$, slow feature analysis seeks to find a mapping of features $f_\theta(x_i) \in R^m$ such that
    
    \begin{equation}
    \theta = \text{argmin}_\theta ||f_\theta(x_i) - f_\theta(x_{i-1})||_2
    \end{equation}
    subject to
    \begin{equation}
    E[f(x_i)_j \cdot f(x_i)_k] - E[f(x_i)_j]\cdot E[f(x_i)_k] = \mathbf{1}\{j = k\}
    \label{eq:var_const}
    \end{equation}
    where the variance constraint in equation \ref{eq:var_const} is important to avoid the degenerate solution of $f(x) = 0$ and to encourage the learned features to be de-correlated. Deep networks have been used to learn slow feautures in the past \cite{sun2014dl}, but to our knowledge our work is the first to do so in a fully probabilistic way. 
    
    A probabilistic reformulation of slow feature analysis presented in \cite{turner2007maximum} showed that slow feature analysis can be motivated by placing a gaussian random walk prior on the model's latent variables as
    \begin{equation}
        p(f(x_i) | f(x_{i-1})) = \text{Normal}(\lambda \cdot f(x_{i-1}), \sigma^2 \cdot I).
    \end{equation}
    
    The prior on our model's temporally varying features is very similar to this formulation as can be seen in line 2 of equation \ref{eq:temporal_prior_gen}. In our model, we simply remove the damping factor $\lambda$. 

\subsection{Variational auto-encoders}
    Variational auto-encoders, originally presented in \cite{kingma2013auto}, are a probabilistic extension of auto-encoders. They reformulate the auto-encoder model as a variational inference problem. In this framework we assume that latent features $h$ are drawn from some prior distribution $p(h)$ and the data $x$ are then drawn from some distribution $p_\theta(x | h)$. Due to the intractability of computing the posterior $p(h | x)$, we seek to find a variational approximation $q_\phi(z | x)$. We do this by finding the values of $\phi$ and $\theta$ that maximize the log-likelihood of observing the data, $\log p(x_1, \ldots, x_N)$. This is also intractable to optimize, but we can instead maximize a lower bound on the log-likelihood
    \begin{equation}
        L(\phi, \theta, x) = -D_{KL}(q_\phi(h | x) || p(h)) + E_{q(h|x)}[p_\theta(x | h)]
        \label{eq:lower_bound}
    \end{equation}
    
    This can be interpreted as optimizing a reconstruction term $p_\theta(x | h)$ and minimizing the KL-Divergence between our variational approximation to the posterior, $q_\phi(h|x)$, and the prior $p(h)$.
    
    Traditionally $p(h) = \text{Normal}(0, I)$ and $q_\phi(h|x)$ is modeled with a neural network that outputs the $\mu_\phi(x)$ and $\sigma_\phi^2(x)$ such that $q_\phi(h|x) = \text{Normal}(h|\mu_\phi(x), \text{Diag}(\sigma^2_\phi(x)))$. The form of the decoder $p_\theta(x | h)$ depends on the type of data that is being modeled. When dealing with image data, a transpose-convolutional neural network \cite{zeiler2010deconvolutional} is typically used and the pixel intensities are typically modeled as bernoulli random variables.

\subsection{Learning Disentangled Representations with Neural Networks}
    Recent  work has demonstrated that neural networks are capable of learning separable features that correspond to distinct factors of variation in the input data. \cite{kulkarni2015deep} presented a modified training procedure for variational auto-encoders that is capable of achieving this goal. This model requires the ability to sample batches from the dataset that have specific latent features held fixed across the batch. For example, when learning a model of human faces, it must be possible to sample a batch of faces oriented in the same way but with different identities and lighting conditions. In most supervised and unsupervised learning environments, this is not possible.
    
    \cite{higgins2016early} demonstrated that under specific conditions separable features can be learned with a variational auto-encoder simply by re-weighting the loss to more heavily penalize KL-Divergence from the prior. They modify the variational lower-bound seen in equation \ref{eq:lower_bound} to 
    \begin{equation}
        L(\phi, \theta, x) = -\beta D_{KL}(q_\phi(h | x) || p(h)) + E_{q(h|x)}[p_\theta(x | h)]
        \label{eq:modified_lower_bound}
    \end{equation}
    where $\beta$ is a hyper-parameter and is typically chosen to be greater than 1. Given a dataset that is dense in its latent factors of variation and a prior $p(h)$ with independent features, this simple modification has been shown to induce the model's individual features to attend these distinct factors.    

\section{Our Model}
    Our model extends the variational auto-encoder framework to operate on video sequences. Differing from the standard approach, we define a prior over latent frame-features for entire frame sequences, not just individual frames. This prior factors into two parts; information that remains relatively constant throughout the video and information that changes temporally. 
    
    As our model is a variational auto-encoder, we must find parameters $\phi$ and $\theta$ that maximize the lower bound on the log-likelihood given in equation \ref{eq:modified_lower_bound}. We choose equation \ref{eq:modified_lower_bound} over equation \ref{eq:lower_bound} since we are interested in investigating the effect of the strength of the variational regularization on the semantics of our learned representation. In most previous work on variational auto-encoders, each example $x$ is a single image and each latent state $h$ is a vector. In our work, each example $x = [x_1, \ldots, x_N]$ is a sequence of frames and each latent state $h = [h_1, \ldots, h_N]$ is a sequence of latent state vectors corresponding to each frame. Thus, the meaning and interpretation of each term in equation \ref{eq:modified_lower_bound} is different in our model. We motivate and derive each term below.

\subsection{Prior}


    
    We are interested in simple assumptions that can be encoded into the prior that will change the semantics of its representation. For example, \cite{higgins2016early} focused on statistical independence of features. This was encoded using a gaussian prior with diagonal covariance. Given that we are working with time-series data, we focus on the rate of change of the latent features over time.
    
    We begin with the assumption that within short video sequences, high level semantic information can be factored into two discrete sets; information that remains static throughout the sequence and information that changes smoothly throughout the sequence. We also assume that these two factors are independent of one another. Thus, each frame latent $h_i = [h_i^s; h_i^t]$ where $h_i^t$ is the temporally changing factor of the representation and $h_i^s$ is the static factor of the representation. If we define $h^s = [h^s_1, \ldots, h^s_N]$ to the sequence of static latent features, and $h^t = [h^t_1, \ldots, h^t_N]$ to be the sequence of temporally changing latent features, we then obtain
    \begin{equation}
        p(h) = p(h^s, h^t) = p(h^s)\cdot p(h^t).
    \end{equation}
    
    We are left to define $p(h^s)$ and $p(h^t)$. In optimizing a variational auto-encoder, we are interested in priors where there exists an analytic solution to the KL-Divergence between the $q_\phi(h|x)$ and $p(h)$. For this reason, we define $p(h^s)$ and $p(h^t)$ to be hierarchical gaussians.
    
    We would like each $h_i^s$ to be near one another within a single sequence, but between sequences we have made no such assumption. For each video, we sample a global $h^s_0$ (inducing variation between videos) and then sample each $h^s_i$ from a distribution centered around $h^s_0$ giving
    \begin{align}
        p(h^s_0)   &= \text{Normal}(h^s_0 | 0, I)\nonumber\\
        p(h^s_i) &= \text{Normal}(h^s_i | h^s_0, \sigma^2_s\cdot I)
        \label{eq:static_prior_gen}
    \end{align}
    where $\sigma^2_s$ is a hyper-parameter which should be less than 1 to give the desired results. This prior produces normally distributed clusters of features.
    
    We would like each $h^t_i$ and $h^t_j$ to be near one another when $i$ is near $j$ to encourage smoothness but otherwise wish to encourage consistent movement. For this reason, we define $p(h^t)$ as a first-order gaussian random walk giving 
    \begin{align}
        p(h^t_1) &= \text{Normal}(h^t_1 | 0, I)\nonumber\\
        p(h^t_i) &= \text{Normal}(h^t_i | h^t_{i-1}, \sigma^2_t\cdot I)
        \label{eq:temporal_prior_gen}
    \end{align}
    where $\sigma^2_t$ is a hyper-parameter which should be less than 1 to give the desired results.
    
\subsection{Prior PDF}
    Crucial to deriving the KL-Divergence term in equation \ref{eq:modified_lower_bound}, we must compute the PDF of each factor in the prior. In the equations presented below we define $N$ as the number of frames and $F$ as the number of features in each factor.
    
    The PDF of the prior over $h^t$ can be computed simply as
    
    \begin{equation}   
        p(h^t) = p(h^t_1, \ldots, h^t_N) = \text{Normal}(h^t_1| 0, I) \cdot \prod_{i=2}^N \text{Normal}(h^t_i| h^t_{i-1}, \sigma^2_t \cdot I)
    \end{equation}
    given the first order markov chain assumption.
    
    The PDF of the prior over $h^s$ can be computed as 
    \begin{align}
       p(h^s) = p(h^s_1, &\ldots, h^s_N) =\nonumber\\
       &C^F \cdot \exp\left(-\frac{1}{2} \sum_{j=1}^F \left(\frac{(\widehat{h_s^2})_j - (\widehat{h_s})^2_j}{\sigma^2_s/N} + \frac{(\widehat{h_s})_j^2}{\sigma^2_s/N + 1}\right)\right)
       \label{eq:static_prior}
    \end{align}
    where
    \begin{align}
       C &= \left(\frac{1}{2\pi}\right)^\frac{N-1}{2} \cdot \left(\frac{\sigma^2_s/N}{(\sigma^2_s)^N}\right)^\frac{1}{2} \cdot \left(\frac{1}{2\pi(\sigma^2_s + 1)}\right)^{\frac{1}{2}}\nonumber\\
       \widehat{h^2_s} &= \sum_{i=1}^N \frac{(h^s_i)^2}{N}\nonumber\\
       \widehat{h_s} &= \sum_{i=1}^N \frac{h^s_i}{N}.
    \end{align}
    
    The static prior (equation \ref{eq:static_prior}) can be better understood by viewing that likely sequences will have feature-wise variance near $\sigma^2_s$ and have mean near 0.

\subsection{Variational Approximation to Posterior}
    Our model seeks to use only its prior $p(h)$ to learn a factored representation. For that reason, we limit the expressive power of $q_\phi(h|x)$. Under this variational approximation, each frame's latent features $h_i$ are independent given the others. This leads to
    \begin{align}
        q_\phi(h | x) &= q_\phi(h_1,\ldots, h_N | x_1, \ldots, x_N)\nonumber\\
        &= \prod_{i=1}^N q_\phi(h_i | x_i)\nonumber\\
        &= \prod_{i=1}^N\text{Normal}(\mu_\phi(x_i), \sigma^2_\phi(x_i)).
    \end{align}
    where $\mu_\phi$ and $\sigma^2_\phi$ are deterministic functions of a single frame input, parametrized by $\phi$.  
 
\subsection{KL Divergence Term}
    
    The KL-Divergence term in equation \ref{eq:modified_lower_bound} can be approximated via a monte-carlo simulation, but that is known \cite{kingma2013auto} to have high variance which can make training difficult. Under the defined model, there exists and analytic solution which we present here. For ease of notation, we ignore the parameter $\phi$ in the following equations. 
    
    The KL-Divergence can be broken up as below.
    \begin{align}
        D_{KL}(q(h| x) || p(h)) = &E_{q(h|x)}[\log q(h|x)] - E_{q(h|x)}[\log p(h^s)] \nonumber\\
        &- E_{q(h|x)}[\log p(h^t)]\nonumber\\
    \end{align}
    We expand each term. The entropy term (equation \ref{eq:entropy_term})
    \begin{align}
        E_{q(h|x)}[\log q(h|x)] &= &&\sum_{i=1}^N E_{q(h_i|x_i)}[\log q(h_i|x_i)] \nonumber\\
        &= &&-\frac{NF}{2} \log(2\pi)\nonumber\\
        & &&- \frac{1}{2}\sum_{i=1}^N\sum_{j=1}^F (1 + \log(\sigma^2(x_i)_j))\nonumber\\
        \label{eq:entropy_term}
    \end{align}
    is similar to that of \cite{kingma2013auto}, but instead of a single frame, we are computing the entropy of the entire sequence of latent features.
    
    The temporal term (equation \ref{eq:temporal_term})
    \begin{align}
        E_{q(h|x)}[\log p(h_t|x)] = &-\frac{F}{2}\log(2\pi)\nonumber\\
        &- \frac{1}{2}\sum_{j=1}^F (\mu(x_1)_j^2 + \sigma^2(x_1)_j)\nonumber\\
        &-\frac{(N-1)F}{2}\log(2\pi \sigma^2_t)\nonumber\\
        &- \frac{1}{2\sigma^2_t} \sum_{j=1}^F\sum_{i=2}^N (\mu(x_i)_j - \mu(x_{i-1})_j)^2 \nonumber\\
        &- \frac{1}{2\sigma^2_t}\sum_{j=1}^F\sum_{i=2}^N(\sigma^2(x_i)_j) + \sigma^2(x_{i-1})_j)
        \label{eq:temporal_term}
    \end{align}
    is similar to the cross-entropy term of \cite{kingma2013auto}. We are representing $p(h^t)$ in terms of its time differences $h^t_i - h^t_{i-1}$ which are themselves gaussian with means $\mu(x_i) - \mu(x_{i-1})$ and variance $\sigma^2(x_i) + \sigma^2(x_{i-1})$. Thus, we are computing the cross-entropy between these time-delta distributions and normal distributions with mean 0 and variance $\sigma^2_t$.
    
    The static term (equation \ref{eq:static_term})
    \begin{align}
        E_{q(h|x)}[\log p(&h^s)] =\nonumber\\
        &-\frac{N}{2\sigma^2_s}\sum_{j=1}^F E_{q(h|x)}[(\widehat{h_s^2})_j]\nonumber\\
        &-\left(\frac{N}{2(\sigma^2_s + N)} - \frac{N}{2\sigma^2_s}\right) \sum_{j=1}^F E_{q(h|x)}[(\widehat{h_s})_j^2]\nonumber\\
        &+ F\cdot\log C\nonumber\\
        \label{eq:static_term}
    \end{align}
    where
    \begin{equation}
         E_{q(h|x)}[(\widehat{h_s^2})_j] =  \sum_{i=1}^N \frac{\mu(x_i)_j^2 + \sigma^2(x_i)_j}{N}
    \end{equation}
    and
    \begin{equation}
         E_{q(h|x)}[(\widehat{h_s})_j^2] = \frac{E_{q(h|x)}[(\widehat{h_s^2})_j]}{N} + \frac{2}{N^2}\sum_{i=1}^{N-1} \mu(x_i)_j \sum_{k=i+1}^N \mu(x_k)_j
    \end{equation}
    can be viewed penalizing a feature sequence $h^s$ who's mean is far from 0 and who's variance is far from $\sigma^2_s$.
    
    The full KL-Divergence can be computed by adding these terms. While verbose, the KL-Divergence term is easy to compute, has well conditioned gradients, and is simple to minimize. We initially experimented with estimating the KL-Divergence via a monte-carlo simulation but found the variance of the estimate to be far too high to optimize. 


\section{Experiments}
\begin{figure}
    \begin{center}
        \begin{subfigure}[b]{.05\textwidth}
                \includegraphics[width=\linewidth]{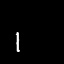}
        \end{subfigure}%
        \hspace{.001\textwidth}
        \begin{subfigure}[b]{.05\textwidth}
                \includegraphics[width=\linewidth]{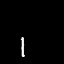}
        \end{subfigure}%
        \hspace{.001\textwidth}
        \begin{subfigure}[b]{.05\textwidth}
                \includegraphics[width=\linewidth]{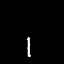}
        \end{subfigure}%
        \hspace{.001\textwidth}
        \begin{subfigure}[b]{.05\textwidth}
                \includegraphics[width=\linewidth]{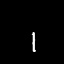}
        \end{subfigure}%
        \hspace{.001\textwidth}
        \begin{subfigure}[b]{.05\textwidth}
                \includegraphics[width=\linewidth]{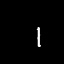}
        \end{subfigure}%
        \hspace{.001\textwidth}
        \begin{subfigure}[b]{.05\textwidth}
                \includegraphics[width=\linewidth]{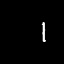}
        \end{subfigure}%
        \hspace{.001\textwidth}
        \begin{subfigure}[b]{.05\textwidth}
                \includegraphics[width=\linewidth]{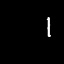}
        \end{subfigure}%
        \hspace{.001\textwidth}
        \begin{subfigure}[b]{.05\textwidth}
                \includegraphics[width=\linewidth]{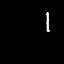}
        \end{subfigure}%
    \end{center}
    \begin{center}
        \begin{subfigure}[b]{.05\textwidth}
                \includegraphics[width=\linewidth]{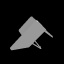}
        \end{subfigure}%
        \hspace{.001\textwidth}
        \begin{subfigure}[b]{.05\textwidth}
                \includegraphics[width=\linewidth]{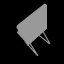}
        \end{subfigure}%
        \hspace{.001\textwidth}
        \begin{subfigure}[b]{.05\textwidth}
                \includegraphics[width=\linewidth]{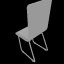}
        \end{subfigure}%
        \hspace{.001\textwidth}
        \begin{subfigure}[b]{.05\textwidth}
                \includegraphics[width=\linewidth]{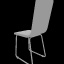}
        \end{subfigure}%
        \hspace{.001\textwidth}
        \begin{subfigure}[b]{.05\textwidth}
                \includegraphics[width=\linewidth]{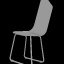}
        \end{subfigure}%
        \hspace{.001\textwidth}
        \begin{subfigure}[b]{.05\textwidth}
                \includegraphics[width=\linewidth]{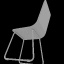}
        \end{subfigure}%
        \hspace{.001\textwidth}
        \begin{subfigure}[b]{.05\textwidth}
                \includegraphics[width=\linewidth]{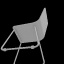}
        \end{subfigure}%
        \hspace{.001\textwidth}
        \begin{subfigure}[b]{.05\textwidth}
                \includegraphics[width=\linewidth]{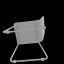}
        \end{subfigure}%
        \caption{Example videos from our 2 datasets.}
        \label{fig:example_data}
    \end{center}
    \end{figure}

   In our experiments it is important to have a quantitative metric for determining the success of our model to disentangle static and temporal information. For in-the-wild video data, it is difficult to come up with such a metric. For that reason, we have restricted our experimentation to datasets where there exists a clean, high-level distinction between temporal and static information. We use two artificially generated datasets to test the success of our algorithm: the bouncing MNIST \cite{srivastava2015unsupervised} dataset and a dataset of rotating chair models taken from the ShapeNet dataset \cite{chang2015shapenet}. Example sequences from these datasets can be seen in figure \ref{fig:example_data}. The information is factored as such:
   
   \begin{enumerate}
        \item Bouncing MNIST
        \begin{labeling}{}
        \item [Static:] Which character is moving
        \item [Temporal:] The location of the character
        \end{labeling}
        
        \item Rotating Chairs
        \begin{labeling}{}
        \item [Static:] Which chair is rotating
        \item [Temporal:] The orientation of the chair
        \end{labeling}
   
   \end{enumerate}
   
   We compare our approach with two baselines; a standard variational auto-encoder that operates on individual frames and a probabilistic variant of slow feature analysis. This variant is similar to our model but the prior on all of the latent features is a gaussian markov random walk. This prior is identical to the prior on the temporal factor of our model (equation \ref{eq:temporal_prior_gen}).

\subsection{Quantifying Disentanglement}
    We present a method to quantify our model's success in disentangling its representation. We train simple linear classifiers \cite{higgins2016early} on subsets of our representation to classify inputs based on their latent factors of variation. We then compute the geometric mean of their accuracies and let a model's score be the max of this value over all divisions of the representation's variables.
    
    Given an encoder $E$, we encode our dataset $H = E(D) \in R^{N \times d}$ where $N$ is the size of the dataset and $d$ is the number of features in the representation. We split the features into two datasets $H_s$ and $H_t$ and train $l2$-svm classifiers for two classification targets $y_s$ and $y_t$ on both datasets.
    
    
    \begin{equation}
    a_s(S) = \frac{\text{Accuracy}(S, y_s)}{\text{Accuracy}(S, y_t)} \nonumber\\
    \end{equation}
    \begin{equation}
    a_t(S) = \frac{\text{Accuracy}(S, y_t)}{\text{Accuracy}(S, y_s)} \nonumber\\
    \end{equation}
    \begin{equation}
    \text{score} = \max_{H_s, H_t \in H} \sqrt{a_s(H_s) a_t(H_t)}
    \label{eq:d_score}
    \end{equation}
    
    In our model, it is intuitive which features should constitute $H_s$ and $H_t$ and we use these to compute the disentanglement score. In our benchmarks, we compute the score as the maximum over all possible combinations of features as shown in equation \ref{eq:d_score}.
    
    Intuitively, $a_s$ should be high for subsets of features that correspond only to static information and $a_t$ should be high for subsets of features that correspond to only temporal information.
    
    For each dataset tested we devise a classification task for each set of features. They are as follows:
    
    \begin{enumerate}
        \item Bouncing MNIST
        \begin{labeling}{}
        \item [Static:] Classify the digit
        \item [Temporal:] Classify the location of the digit (bucketed into a 3$\times$3 grid)
        \end{labeling}
        
        \item Rotating Chairs
        \begin{labeling}{}
        \item [Static:] Classify which chair is in the frame (out of a held-out set of 10 chairs not used in training)
        \item [Temporal:] Classify if a sequence of 3 chair frames are temporally ordered (1-2-3 and 3-2-1 are true and 1-3-2 and 2-1-3 are false)
        \end{labeling}
    \end{enumerate}
    
    A model which learns to factor its representation correctly will have one factor which performs well on each task and poorly on the other, giving a high value for the disentanglement score defined in equation \ref{eq:d_score}.
    
\subsection{Experimental Setup}
    We train the video models for 30,000 iterations with a batch size of 64 videos. Each video has 16 frames. We use the Adam \cite{kingma2014adam} learning rule. We use a learning rate of .001 multiplied by .1 every 10,000 iterations. The VAE benchmark was trained with a batch size of 1024 to ensure that the images per batch are consistent.
    
    We trained versions our model on the Bouncing MNIST dataset with 2 and 4 features in each factor. We compare against the benchmark models trained with 2 and 4 features in total. On the rotating chairs dataset we trained our model with 4 features in each factor and compare against the benchmark models trained with 4 and 8 features in total.
    
    Tables \ref{tab:enc_arch} and \ref{tab:dec_arch} outline our chosen model architectures. For the encoder, we use a convolutional architecture with batch normalization applied before the nonlinearity at every layer except for the model output layers. We use the rectified-linear nonlinearity for all layers.

    In the decoder, the output of our second linear layer is reshaped into a $4\times4\times128$ tensor. All subsequent layers are transpose convolutional layers in the form of $$f(x) = \text{relu}(W*x + b)$$ where $*$ indicates a transpose-convolution operation \cite{zeiler2010deconvolutional}. We use batch normalization in all decoder layers except for the output layer.
    
    This is the first and only model architecture we tested. We believe better results could be obtained through experimenting with the model architecture, but that is beyond the scope of this work. In all experiments we set $\sigma^2_s = \sigma^2_t = 0.01$. This was the only value of these parameters tested. We believe better results could also be obtained by optimizing this parameter, but we leave this for further work as well. 
    
    \cite{higgins2016early} demonstrated that the strength of the distributional regularization term in the VAE lowerbound can greatly impact the semantics of the learned features. Because of this, we run all models (ours and benchmarks) with a regularization strength of 1 (bayes solution), 2, and 4.
        
\subsection{Qualitative Results}    
    We present qualitative results that visually demonstrate our model's ability to factor its representation between the static and time-varying features in the data.
    
\begin{figure}
    \begin{center}
        \begin{subfigure}[b]{.1\textwidth}
                \includegraphics[width=\linewidth]{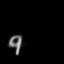}
        \end{subfigure}%
        \hspace{.001\textwidth}
        \begin{subfigure}[b]{.1\textwidth}
                \includegraphics[width=\linewidth]{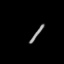}
        \end{subfigure}%

        \begin{subfigure}[b]{.1\textwidth}
                \includegraphics[width=\linewidth]{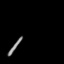}
        \end{subfigure}%
        \hspace{.001\textwidth}
        \begin{subfigure}[b]{.1\textwidth}
                \includegraphics[width=\linewidth]{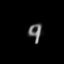}
        \end{subfigure}%
        \caption{The top line contains two decoded feature samples. The bottom line shows those same sets of feature fed through the decoder with their static and temporal components swapped.}
        \label{fig:factor_swap}
    \end{center}
    \end{figure}

    In figure \ref{fig:factor_swap}, we show two frame embeddings drawn from the prior and pass them through the decoder of a model trained on the Bouncing MNIST dataset. We then swap the static factors of their embeddings and pass them through the decoder again. It can be seen that in each image, the character identities swap, but their locations remain the same, demonstrating the different semantic interpretations of each factor of features. 
    
\begin{figure}
    \begin{center}
        \begin{subfigure}[b]{.05\textwidth}
                \includegraphics[width=\linewidth]{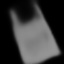}
        \end{subfigure}%
        \hspace{.001\textwidth}
        \begin{subfigure}[b]{.05\textwidth}
                \includegraphics[width=\linewidth]{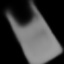}
        \end{subfigure}%
        \hspace{.001\textwidth}
        \begin{subfigure}[b]{.05\textwidth}
                \includegraphics[width=\linewidth]{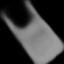}
        \end{subfigure}%
        \hspace{.001\textwidth}
        \begin{subfigure}[b]{.05\textwidth}
                \includegraphics[width=\linewidth]{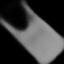}
        \end{subfigure}%
        \hspace{.001\textwidth}
        \begin{subfigure}[b]{.05\textwidth}
                \includegraphics[width=\linewidth]{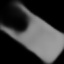}
        \end{subfigure}%
        \hspace{.001\textwidth}
        \begin{subfigure}[b]{.05\textwidth}
                \includegraphics[width=\linewidth]{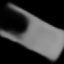}
        \end{subfigure}%
        \hspace{.001\textwidth}
        \begin{subfigure}[b]{.05\textwidth}
                \includegraphics[width=\linewidth]{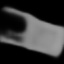}
        \end{subfigure}%
        \hspace{.001\textwidth}
        \begin{subfigure}[b]{.05\textwidth}
                \includegraphics[width=\linewidth]{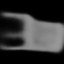}
        \end{subfigure}%
    \end{center}
    \begin{center}
        \begin{subfigure}[b]{.05\textwidth}
                \includegraphics[width=\linewidth]{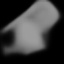}
        \end{subfigure}%
        \hspace{.001\textwidth}
        \begin{subfigure}[b]{.05\textwidth}
                \includegraphics[width=\linewidth]{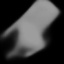}
        \end{subfigure}%
        \hspace{.001\textwidth}
        \begin{subfigure}[b]{.05\textwidth}
                \includegraphics[width=\linewidth]{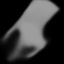}
        \end{subfigure}%
        \hspace{.001\textwidth}
        \begin{subfigure}[b]{.05\textwidth}
                \includegraphics[width=\linewidth]{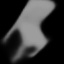}
        \end{subfigure}%
        \hspace{.001\textwidth}
        \begin{subfigure}[b]{.05\textwidth}
                \includegraphics[width=\linewidth]{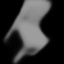}
        \end{subfigure}%
        \hspace{.001\textwidth}
        \begin{subfigure}[b]{.05\textwidth}
                \includegraphics[width=\linewidth]{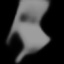}
        \end{subfigure}%
        \hspace{.001\textwidth}
        \begin{subfigure}[b]{.05\textwidth}
                \includegraphics[width=\linewidth]{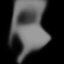}
        \end{subfigure}%
        \hspace{.001\textwidth}
        \begin{subfigure}[b]{.05\textwidth}
                \includegraphics[width=\linewidth]{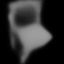}
        \end{subfigure}%
        \caption{The same sequence of temporal encodings with two different static encodings. The static encodings are constant across each row and the temporal encodings are constant across each column.}
        \label{fig:factor_swap_chair}
    \end{center}
    \end{figure}

    In figure \ref{fig:factor_swap_chair}, we show two sequences of frames. The images in each row are generated using the same sequence of temporal factor encodings but have two different fixed static factor encodings. Both sequences show chairs rotating the same angle along roughly the same axis, but the chair's appearance and starting orientations are different. This indicates that the model has learned to represent rotations of a given angle around an axis as a sequence of these temporal factor encodings while representing little information about the chair's identity. 
    
\begin{figure}
    \begin{center}
        \begin{subfigure}[b]{.05\textwidth}
                \includegraphics[width=\linewidth]{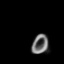}
        \end{subfigure}%
        \hspace{.001\textwidth}
        \begin{subfigure}[b]{.05\textwidth}
                \includegraphics[width=\linewidth]{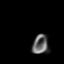}
        \end{subfigure}%
        \hspace{.001\textwidth}
        \begin{subfigure}[b]{.05\textwidth}
                \includegraphics[width=\linewidth]{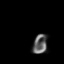}
        \end{subfigure}%
        \hspace{.001\textwidth}
        \begin{subfigure}[b]{.05\textwidth}
                \includegraphics[width=\linewidth]{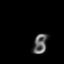}
        \end{subfigure}%
        \hspace{.001\textwidth}
        \begin{subfigure}[b]{.05\textwidth}
                \includegraphics[width=\linewidth]{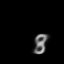}
        \end{subfigure}%
        \hspace{.001\textwidth}
        \begin{subfigure}[b]{.05\textwidth}
                \includegraphics[width=\linewidth]{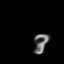}
        \end{subfigure}%
        \hspace{.001\textwidth}
        \begin{subfigure}[b]{.05\textwidth}
                \includegraphics[width=\linewidth]{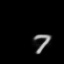}
        \end{subfigure}%
        \hspace{.001\textwidth}
        \begin{subfigure}[b]{.05\textwidth}
                \includegraphics[width=\linewidth]{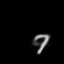}
        \end{subfigure}%
    \end{center}
    \begin{center}
        \begin{subfigure}[b]{.05\textwidth}
                \includegraphics[width=\linewidth]{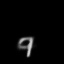}
        \end{subfigure}%
        \hspace{.001\textwidth}
        \begin{subfigure}[b]{.05\textwidth}
                \includegraphics[width=\linewidth]{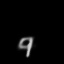}
        \end{subfigure}%
        \hspace{.001\textwidth}
        \begin{subfigure}[b]{.05\textwidth}
                \includegraphics[width=\linewidth]{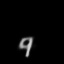}
        \end{subfigure}%
        \hspace{.001\textwidth}
        \begin{subfigure}[b]{.05\textwidth}
                \includegraphics[width=\linewidth]{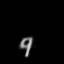}
        \end{subfigure}%
        \hspace{.001\textwidth}
        \begin{subfigure}[b]{.05\textwidth}
                \includegraphics[width=\linewidth]{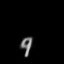}
        \end{subfigure}%
        \hspace{.001\textwidth}
        \begin{subfigure}[b]{.05\textwidth}
                \includegraphics[width=\linewidth]{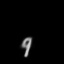}
        \end{subfigure}%
        \hspace{.001\textwidth}
        \begin{subfigure}[b]{.05\textwidth}
                \includegraphics[width=\linewidth]{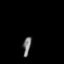}
        \end{subfigure}%
        \hspace{.001\textwidth}
        \begin{subfigure}[b]{.05\textwidth}
                \includegraphics[width=\linewidth]{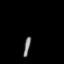}
        \end{subfigure}%
        \hspace{.001\textwidth}
    \end{center}
    \begin{center}
        \begin{subfigure}[b]{.05\textwidth}
                \includegraphics[width=\linewidth]{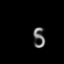}
        \end{subfigure}%
        \hspace{.001\textwidth}
        \begin{subfigure}[b]{.05\textwidth}
                \includegraphics[width=\linewidth]{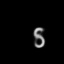}
        \end{subfigure}%
        \hspace{.001\textwidth}
        \begin{subfigure}[b]{.05\textwidth}
                \includegraphics[width=\linewidth]{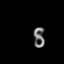}
        \end{subfigure}%
        \hspace{.001\textwidth}
        \begin{subfigure}[b]{.05\textwidth}
                \includegraphics[width=\linewidth]{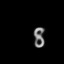}
        \end{subfigure}%
        \hspace{.001\textwidth}
        \begin{subfigure}[b]{.05\textwidth}
                \includegraphics[width=\linewidth]{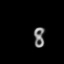}
        \end{subfigure}%
        \hspace{.001\textwidth}
        \begin{subfigure}[b]{.05\textwidth}
                \includegraphics[width=\linewidth]{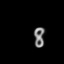}
        \end{subfigure}%
        \hspace{.001\textwidth}
        \begin{subfigure}[b]{.05\textwidth}
                \includegraphics[width=\linewidth]{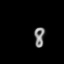}
        \end{subfigure}%
        \hspace{.001\textwidth}
        \begin{subfigure}[b]{.05\textwidth}
                \includegraphics[width=\linewidth]{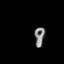}
        \end{subfigure}%
        \hspace{.001\textwidth}
    \end{center}
    \begin{center}
        \begin{subfigure}[b]{.05\textwidth}
                \includegraphics[width=\linewidth]{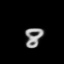}
        \end{subfigure}%
        \hspace{.001\textwidth}
        \begin{subfigure}[b]{.05\textwidth}
                \includegraphics[width=\linewidth]{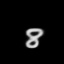}
        \end{subfigure}%
        \hspace{.001\textwidth}
        \begin{subfigure}[b]{.05\textwidth}
                \includegraphics[width=\linewidth]{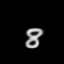}
        \end{subfigure}%
        \hspace{.001\textwidth}
        \begin{subfigure}[b]{.05\textwidth}
                \includegraphics[width=\linewidth]{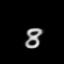}
        \end{subfigure}%
        \hspace{.001\textwidth}
        \begin{subfigure}[b]{.05\textwidth}
                \includegraphics[width=\linewidth]{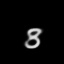}
        \end{subfigure}%
        \hspace{.001\textwidth}
        \begin{subfigure}[b]{.05\textwidth}
                \includegraphics[width=\linewidth]{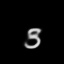}
        \end{subfigure}%
        \hspace{.001\textwidth}
        \begin{subfigure}[b]{.05\textwidth}
                \includegraphics[width=\linewidth]{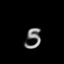}
        \end{subfigure}%
        \hspace{.001\textwidth}
        \begin{subfigure}[b]{.05\textwidth}
                \includegraphics[width=\linewidth]{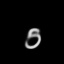}
        \end{subfigure}%
        \hspace{.001\textwidth}
    \end{center}
    \begin{center}
        \begin{subfigure}[b]{.05\textwidth}
                \includegraphics[width=\linewidth]{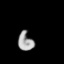}
        \end{subfigure}%
        \hspace{.001\textwidth}
        \begin{subfigure}[b]{.05\textwidth}
                \includegraphics[width=\linewidth]{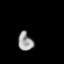}
        \end{subfigure}%
        \hspace{.001\textwidth}
        \begin{subfigure}[b]{.05\textwidth}
                \includegraphics[width=\linewidth]{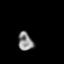}
        \end{subfigure}%
        \hspace{.001\textwidth}
        \begin{subfigure}[b]{.05\textwidth}
                \includegraphics[width=\linewidth]{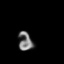}
        \end{subfigure}%
        \hspace{.001\textwidth}
        \begin{subfigure}[b]{.05\textwidth}
                \includegraphics[width=\linewidth]{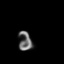}
        \end{subfigure}%
        \hspace{.001\textwidth}
        \begin{subfigure}[b]{.05\textwidth}
                \includegraphics[width=\linewidth]{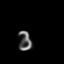}
        \end{subfigure}%
        \hspace{.001\textwidth}
        \begin{subfigure}[b]{.05\textwidth}
                \includegraphics[width=\linewidth]{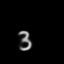}
        \end{subfigure}%
        \hspace{.001\textwidth}
        \begin{subfigure}[b]{.05\textwidth}
                \includegraphics[width=\linewidth]{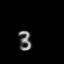}
        \end{subfigure}%
        \hspace{.001\textwidth}
    \end{center}
    \begin{center}
        \begin{subfigure}[b]{.05\textwidth}
                \includegraphics[width=\linewidth]{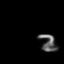}
        \end{subfigure}%
        \hspace{.001\textwidth}
        \begin{subfigure}[b]{.05\textwidth}
                \includegraphics[width=\linewidth]{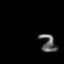}
        \end{subfigure}%
        \hspace{.001\textwidth}
        \begin{subfigure}[b]{.05\textwidth}
                \includegraphics[width=\linewidth]{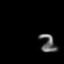}
        \end{subfigure}%
        \hspace{.001\textwidth}
        \begin{subfigure}[b]{.05\textwidth}
                \includegraphics[width=\linewidth]{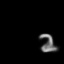}
        \end{subfigure}%
        \hspace{.001\textwidth}
        \begin{subfigure}[b]{.05\textwidth}
                \includegraphics[width=\linewidth]{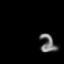}
        \end{subfigure}%
        \hspace{.001\textwidth}
        \begin{subfigure}[b]{.05\textwidth}
                \includegraphics[width=\linewidth]{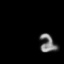}
        \end{subfigure}%
        \hspace{.001\textwidth}
        \begin{subfigure}[b]{.05\textwidth}
                \includegraphics[width=\linewidth]{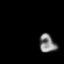}
        \end{subfigure}%
        \hspace{.001\textwidth}
        \begin{subfigure}[b]{.05\textwidth}
                \includegraphics[width=\linewidth]{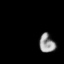}
        \end{subfigure}%
        \hspace{.001\textwidth}
    \end{center}
    \caption{Linear interpolation between two points in static factor with temporal factor held fixed for the Bouncing MNIST dataset.}
    \label{fig:static_interp}
    \end{figure}

\begin{figure}
    \begin{center}
        \begin{subfigure}[b]{.05\textwidth}
                \includegraphics[width=\linewidth]{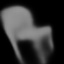}
        \end{subfigure}%
        \hspace{.001\textwidth}
        \begin{subfigure}[b]{.05\textwidth}
                \includegraphics[width=\linewidth]{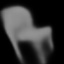}
        \end{subfigure}%
        \hspace{.001\textwidth}
        \begin{subfigure}[b]{.05\textwidth}
                \includegraphics[width=\linewidth]{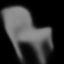}
        \end{subfigure}%
        \hspace{.001\textwidth}
        \begin{subfigure}[b]{.05\textwidth}
                \includegraphics[width=\linewidth]{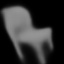}
        \end{subfigure}%
        \hspace{.001\textwidth}
        \begin{subfigure}[b]{.05\textwidth}
                \includegraphics[width=\linewidth]{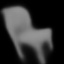}
        \end{subfigure}%
        \hspace{.001\textwidth}
        \begin{subfigure}[b]{.05\textwidth}
                \includegraphics[width=\linewidth]{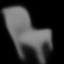}
        \end{subfigure}%
        \hspace{.001\textwidth}
        \begin{subfigure}[b]{.05\textwidth}
                \includegraphics[width=\linewidth]{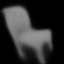}
        \end{subfigure}%
        \hspace{.001\textwidth}
        \begin{subfigure}[b]{.05\textwidth}
                \includegraphics[width=\linewidth]{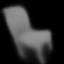}
        \end{subfigure}%
    \end{center}
    \begin{center}
        \begin{subfigure}[b]{.05\textwidth}
                \includegraphics[width=\linewidth]{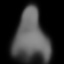}
        \end{subfigure}%
        \hspace{.001\textwidth}
        \begin{subfigure}[b]{.05\textwidth}
                \includegraphics[width=\linewidth]{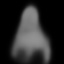}
        \end{subfigure}%
        \hspace{.001\textwidth}
        \begin{subfigure}[b]{.05\textwidth}
                \includegraphics[width=\linewidth]{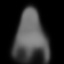}
        \end{subfigure}%
        \hspace{.001\textwidth}
        \begin{subfigure}[b]{.05\textwidth}
                \includegraphics[width=\linewidth]{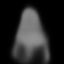}
        \end{subfigure}%
        \hspace{.001\textwidth}
        \begin{subfigure}[b]{.05\textwidth}
                \includegraphics[width=\linewidth]{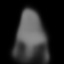}
        \end{subfigure}%
        \hspace{.001\textwidth}
        \begin{subfigure}[b]{.05\textwidth}
                \includegraphics[width=\linewidth]{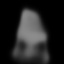}
        \end{subfigure}%
        \hspace{.001\textwidth}
        \begin{subfigure}[b]{.05\textwidth}
                \includegraphics[width=\linewidth]{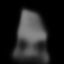}
        \end{subfigure}%
        \hspace{.001\textwidth}
        \begin{subfigure}[b]{.05\textwidth}
                \includegraphics[width=\linewidth]{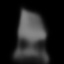}
        \end{subfigure}%
        \hspace{.001\textwidth}
    \end{center}
    \begin{center}
        \begin{subfigure}[b]{.05\textwidth}
                \includegraphics[width=\linewidth]{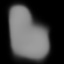}
        \end{subfigure}%
        \hspace{.001\textwidth}
        \begin{subfigure}[b]{.05\textwidth}
                \includegraphics[width=\linewidth]{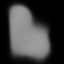}
        \end{subfigure}%
        \hspace{.001\textwidth}
        \begin{subfigure}[b]{.05\textwidth}
                \includegraphics[width=\linewidth]{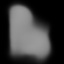}
        \end{subfigure}%
        \hspace{.001\textwidth}
        \begin{subfigure}[b]{.05\textwidth}
                \includegraphics[width=\linewidth]{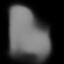}
        \end{subfigure}%
        \hspace{.001\textwidth}
        \begin{subfigure}[b]{.05\textwidth}
                \includegraphics[width=\linewidth]{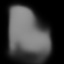}
        \end{subfigure}%
        \hspace{.001\textwidth}
        \begin{subfigure}[b]{.05\textwidth}
                \includegraphics[width=\linewidth]{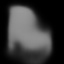}
        \end{subfigure}%
        \hspace{.001\textwidth}
        \begin{subfigure}[b]{.05\textwidth}
                \includegraphics[width=\linewidth]{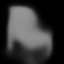}
        \end{subfigure}%
        \hspace{.001\textwidth}
        \begin{subfigure}[b]{.05\textwidth}
                \includegraphics[width=\linewidth]{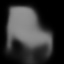}
        \end{subfigure}%
        \hspace{.001\textwidth}
    \end{center}
    \begin{center}
        \begin{subfigure}[b]{.05\textwidth}
                \includegraphics[width=\linewidth]{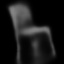}
        \end{subfigure}%
        \hspace{.001\textwidth}
        \begin{subfigure}[b]{.05\textwidth}
                \includegraphics[width=\linewidth]{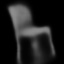}
        \end{subfigure}%
        \hspace{.001\textwidth}
        \begin{subfigure}[b]{.05\textwidth}
                \includegraphics[width=\linewidth]{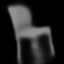}
        \end{subfigure}%
        \hspace{.001\textwidth}
        \begin{subfigure}[b]{.05\textwidth}
                \includegraphics[width=\linewidth]{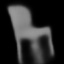}
        \end{subfigure}%
        \hspace{.001\textwidth}
        \begin{subfigure}[b]{.05\textwidth}
                \includegraphics[width=\linewidth]{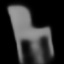}
        \end{subfigure}%
        \hspace{.001\textwidth}
        \begin{subfigure}[b]{.05\textwidth}
                \includegraphics[width=\linewidth]{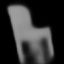}
        \end{subfigure}%
        \hspace{.001\textwidth}
        \begin{subfigure}[b]{.05\textwidth}
                \includegraphics[width=\linewidth]{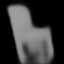}
        \end{subfigure}%
        \hspace{.001\textwidth}
        \begin{subfigure}[b]{.05\textwidth}
                \includegraphics[width=\linewidth]{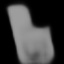}
        \end{subfigure}%
        \hspace{.001\textwidth}
    \end{center}
    \begin{center}
        \begin{subfigure}[b]{.05\textwidth}
                \includegraphics[width=\linewidth]{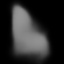}
        \end{subfigure}%
        \hspace{.001\textwidth}
        \begin{subfigure}[b]{.05\textwidth}
                \includegraphics[width=\linewidth]{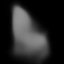}
        \end{subfigure}%
        \hspace{.001\textwidth}
        \begin{subfigure}[b]{.05\textwidth}
                \includegraphics[width=\linewidth]{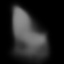}
        \end{subfigure}%
        \hspace{.001\textwidth}
        \begin{subfigure}[b]{.05\textwidth}
                \includegraphics[width=\linewidth]{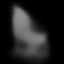}
        \end{subfigure}%
        \hspace{.001\textwidth}
        \begin{subfigure}[b]{.05\textwidth}
                \includegraphics[width=\linewidth]{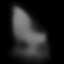}
        \end{subfigure}%
        \hspace{.001\textwidth}
        \begin{subfigure}[b]{.05\textwidth}
                \includegraphics[width=\linewidth]{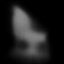}
        \end{subfigure}%
        \hspace{.001\textwidth}
        \begin{subfigure}[b]{.05\textwidth}
                \includegraphics[width=\linewidth]{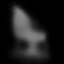}
        \end{subfigure}%
        \hspace{.001\textwidth}
        \begin{subfigure}[b]{.05\textwidth}
                \includegraphics[width=\linewidth]{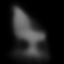}
        \end{subfigure}%
        \hspace{.001\textwidth}
    \end{center}
    \begin{center}
        \begin{subfigure}[b]{.05\textwidth}
                \includegraphics[width=\linewidth]{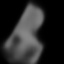}
        \end{subfigure}%
        \hspace{.001\textwidth}
        \begin{subfigure}[b]{.05\textwidth}
                \includegraphics[width=\linewidth]{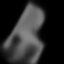}
        \end{subfigure}%
        \hspace{.001\textwidth}
        \begin{subfigure}[b]{.05\textwidth}
                \includegraphics[width=\linewidth]{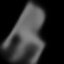}
        \end{subfigure}%
        \hspace{.001\textwidth}
        \begin{subfigure}[b]{.05\textwidth}
                \includegraphics[width=\linewidth]{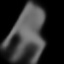}
        \end{subfigure}%
        \hspace{.001\textwidth}
        \begin{subfigure}[b]{.05\textwidth}
                \includegraphics[width=\linewidth]{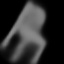}
        \end{subfigure}%
        \hspace{.001\textwidth}
        \begin{subfigure}[b]{.05\textwidth}
                \includegraphics[width=\linewidth]{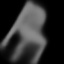}
        \end{subfigure}%
        \hspace{.001\textwidth}
        \begin{subfigure}[b]{.05\textwidth}
                \includegraphics[width=\linewidth]{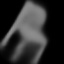}
        \end{subfigure}%
        \hspace{.001\textwidth}
        \begin{subfigure}[b]{.05\textwidth}
                \includegraphics[width=\linewidth]{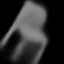}
        \end{subfigure}%
        \hspace{.001\textwidth}
    \end{center}
    \caption{Linear interpolation between two points in static factor with temporal factor held fixed for the Rotating Chairs dataset.}
    \label{fig:static_interp_chair}
    \end{figure}

    In figures \ref{fig:static_interp} and \ref{fig:static_interp_chair}, we present outputs from the decoder of inputs that smoothly interpolate between two sampled points in the static factor while holding a sample from the temporal factor fixed. In the Bouncing MNIST example, the location of the character remains constant but the character's identity smoothly transforms. This demonstrates that the static factor of the embedding encodes no spatial (and therefore temporally-varying) information. In the Rotating Chairs example, the characteristics of the decoded chair change (leg thickness, leg length, leg type, back shape, etc) while the orientation of the chair remains roughly constant. This demonstrates that the static factor of the embedding encodes no orientation information. 
    
\begin{figure}
    \begin{center}
        \begin{subfigure}[b]{.05\textwidth}
                \includegraphics[width=\linewidth]{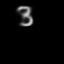}
        \end{subfigure}%
        \hspace{.001\textwidth}
        \begin{subfigure}[b]{.05\textwidth}
                \includegraphics[width=\linewidth]{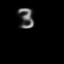}
        \end{subfigure}%
        \hspace{.001\textwidth}
        \begin{subfigure}[b]{.05\textwidth}
                \includegraphics[width=\linewidth]{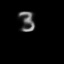}
        \end{subfigure}%
        \hspace{.001\textwidth}
        \begin{subfigure}[b]{.05\textwidth}
                \includegraphics[width=\linewidth]{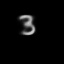}
        \end{subfigure}%
        \hspace{.001\textwidth}
        \begin{subfigure}[b]{.05\textwidth}
                \includegraphics[width=\linewidth]{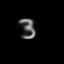}
        \end{subfigure}%
        \hspace{.001\textwidth}
        \begin{subfigure}[b]{.05\textwidth}
                \includegraphics[width=\linewidth]{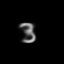}
        \end{subfigure}%
        \hspace{.001\textwidth}
        \begin{subfigure}[b]{.05\textwidth}
                \includegraphics[width=\linewidth]{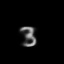}
        \end{subfigure}%
        \hspace{.001\textwidth}
        \begin{subfigure}[b]{.05\textwidth}
                \includegraphics[width=\linewidth]{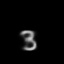}
        \end{subfigure}%
    \end{center}
    \begin{center}
        \begin{subfigure}[b]{.05\textwidth}
                \includegraphics[width=\linewidth]{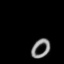}
        \end{subfigure}%
        \hspace{.001\textwidth}
        \begin{subfigure}[b]{.05\textwidth}
                \includegraphics[width=\linewidth]{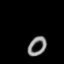}
        \end{subfigure}%
        \hspace{.001\textwidth}
        \begin{subfigure}[b]{.05\textwidth}
                \includegraphics[width=\linewidth]{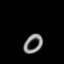}
        \end{subfigure}%
        \hspace{.001\textwidth}
        \begin{subfigure}[b]{.05\textwidth}
                \includegraphics[width=\linewidth]{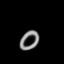}
        \end{subfigure}%
        \hspace{.001\textwidth}
        \begin{subfigure}[b]{.05\textwidth}
                \includegraphics[width=\linewidth]{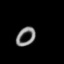}
        \end{subfigure}%
        \hspace{.001\textwidth}
        \begin{subfigure}[b]{.05\textwidth}
                \includegraphics[width=\linewidth]{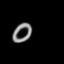}
        \end{subfigure}%
        \hspace{.001\textwidth}
        \begin{subfigure}[b]{.05\textwidth}
                \includegraphics[width=\linewidth]{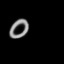}
        \end{subfigure}%
        \hspace{.001\textwidth}
        \begin{subfigure}[b]{.05\textwidth}
                \includegraphics[width=\linewidth]{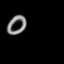}
        \end{subfigure}%
    \end{center}
    \begin{center}
        \begin{subfigure}[b]{.05\textwidth}
                \includegraphics[width=\linewidth]{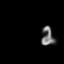}
        \end{subfigure}%
        \hspace{.001\textwidth}
        \begin{subfigure}[b]{.05\textwidth}
                \includegraphics[width=\linewidth]{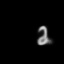}
        \end{subfigure}%
        \hspace{.001\textwidth}
        \begin{subfigure}[b]{.05\textwidth}
                \includegraphics[width=\linewidth]{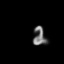}
        \end{subfigure}%
        \hspace{.001\textwidth}
        \begin{subfigure}[b]{.05\textwidth}
                \includegraphics[width=\linewidth]{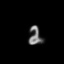}
        \end{subfigure}%
        \hspace{.001\textwidth}
        \begin{subfigure}[b]{.05\textwidth}
                \includegraphics[width=\linewidth]{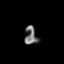}
        \end{subfigure}%
        \hspace{.001\textwidth}
        \begin{subfigure}[b]{.05\textwidth}
                \includegraphics[width=\linewidth]{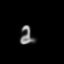}
        \end{subfigure}%
        \hspace{.001\textwidth}
        \begin{subfigure}[b]{.05\textwidth}
                \includegraphics[width=\linewidth]{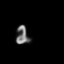}
        \end{subfigure}%
        \hspace{.001\textwidth}
        \begin{subfigure}[b]{.05\textwidth}
                \includegraphics[width=\linewidth]{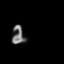}
        \end{subfigure}%
    \end{center}
    \begin{center}
        \begin{subfigure}[b]{.05\textwidth}
                \includegraphics[width=\linewidth]{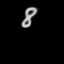}
        \end{subfigure}%
        \hspace{.001\textwidth}
        \begin{subfigure}[b]{.05\textwidth}
                \includegraphics[width=\linewidth]{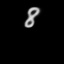}
        \end{subfigure}%
        \hspace{.001\textwidth}
        \begin{subfigure}[b]{.05\textwidth}
                \includegraphics[width=\linewidth]{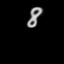}
        \end{subfigure}%
        \hspace{.001\textwidth}
        \begin{subfigure}[b]{.05\textwidth}
                \includegraphics[width=\linewidth]{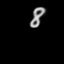}
        \end{subfigure}%
        \hspace{.001\textwidth}
        \begin{subfigure}[b]{.05\textwidth}
                \includegraphics[width=\linewidth]{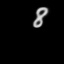}
        \end{subfigure}%
        \hspace{.001\textwidth}
        \begin{subfigure}[b]{.05\textwidth}
                \includegraphics[width=\linewidth]{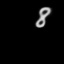}
        \end{subfigure}%
        \hspace{.001\textwidth}
        \begin{subfigure}[b]{.05\textwidth}
                \includegraphics[width=\linewidth]{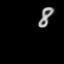}
        \end{subfigure}%
        \hspace{.001\textwidth}
        \begin{subfigure}[b]{.05\textwidth}
                \includegraphics[width=\linewidth]{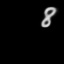}
        \end{subfigure}%
    \end{center}
    \begin{center}
        \begin{subfigure}[b]{.05\textwidth}
                \includegraphics[width=\linewidth]{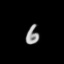}
        \end{subfigure}%
        \hspace{.001\textwidth}
        \begin{subfigure}[b]{.05\textwidth}
                \includegraphics[width=\linewidth]{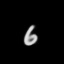}
        \end{subfigure}%
        \hspace{.001\textwidth}
        \begin{subfigure}[b]{.05\textwidth}
                \includegraphics[width=\linewidth]{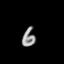}
        \end{subfigure}%
        \hspace{.001\textwidth}
        \begin{subfigure}[b]{.05\textwidth}
                \includegraphics[width=\linewidth]{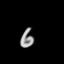}
        \end{subfigure}%
        \hspace{.001\textwidth}
        \begin{subfigure}[b]{.05\textwidth}
                \includegraphics[width=\linewidth]{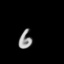}
        \end{subfigure}%
        \hspace{.001\textwidth}
        \begin{subfigure}[b]{.05\textwidth}
                \includegraphics[width=\linewidth]{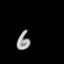}
        \end{subfigure}%
        \hspace{.001\textwidth}
        \begin{subfigure}[b]{.05\textwidth}
                \includegraphics[width=\linewidth]{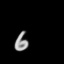}
        \end{subfigure}%
        \hspace{.001\textwidth}
        \begin{subfigure}[b]{.05\textwidth}
                \includegraphics[width=\linewidth]{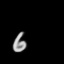}
        \end{subfigure}%
    \end{center}
    \begin{center}
        \begin{subfigure}[b]{.05\textwidth}
                \includegraphics[width=\linewidth]{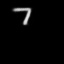}
        \end{subfigure}%
        \hspace{.001\textwidth}
        \begin{subfigure}[b]{.05\textwidth}
                \includegraphics[width=\linewidth]{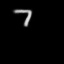}
        \end{subfigure}%
        \hspace{.001\textwidth}
        \begin{subfigure}[b]{.05\textwidth}
                \includegraphics[width=\linewidth]{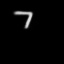}
        \end{subfigure}%
        \hspace{.001\textwidth}
        \begin{subfigure}[b]{.05\textwidth}
                \includegraphics[width=\linewidth]{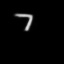}
        \end{subfigure}%
        \hspace{.001\textwidth}
        \begin{subfigure}[b]{.05\textwidth}
                \includegraphics[width=\linewidth]{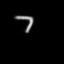}
        \end{subfigure}%
        \hspace{.001\textwidth}
        \begin{subfigure}[b]{.05\textwidth}
                \includegraphics[width=\linewidth]{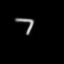}
        \end{subfigure}%
        \hspace{.001\textwidth}
        \begin{subfigure}[b]{.05\textwidth}
                \includegraphics[width=\linewidth]{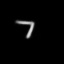}
        \end{subfigure}%
        \hspace{.001\textwidth}
        \begin{subfigure}[b]{.05\textwidth}
                \includegraphics[width=\linewidth]{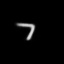}
        \end{subfigure}%
    \end{center}
    \caption{Linear interpolation between two points in temporal factor with static factor held fixed for the Bouncing MNIST dataset.}
    \label{fig:temporal_interp}
    \end{figure}

\begin{figure}
    \begin{center}
        \begin{subfigure}[b]{.05\textwidth}
                \includegraphics[width=\linewidth]{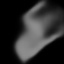}
        \end{subfigure}%
        \hspace{.001\textwidth}
        \begin{subfigure}[b]{.05\textwidth}
                \includegraphics[width=\linewidth]{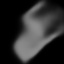}
        \end{subfigure}%
        \hspace{.001\textwidth}
        \begin{subfigure}[b]{.05\textwidth}
                \includegraphics[width=\linewidth]{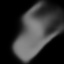}
        \end{subfigure}%
        \hspace{.001\textwidth}
        \begin{subfigure}[b]{.05\textwidth}
                \includegraphics[width=\linewidth]{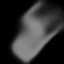}
        \end{subfigure}%
        \hspace{.001\textwidth}
        \begin{subfigure}[b]{.05\textwidth}
                \includegraphics[width=\linewidth]{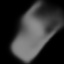}
        \end{subfigure}%
        \hspace{.001\textwidth}
        \begin{subfigure}[b]{.05\textwidth}
                \includegraphics[width=\linewidth]{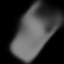}
        \end{subfigure}%
        \hspace{.001\textwidth}
        \begin{subfigure}[b]{.05\textwidth}
                \includegraphics[width=\linewidth]{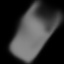}
        \end{subfigure}%
        \hspace{.001\textwidth}
        \begin{subfigure}[b]{.05\textwidth}
                \includegraphics[width=\linewidth]{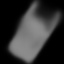}
        \end{subfigure}%
    \end{center}
    \begin{center}
        \begin{subfigure}[b]{.05\textwidth}
                \includegraphics[width=\linewidth]{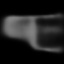}
        \end{subfigure}%
        \hspace{.001\textwidth}
        \begin{subfigure}[b]{.05\textwidth}
                \includegraphics[width=\linewidth]{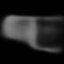}
        \end{subfigure}%
        \hspace{.001\textwidth}
        \begin{subfigure}[b]{.05\textwidth}
                \includegraphics[width=\linewidth]{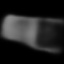}
        \end{subfigure}%
        \hspace{.001\textwidth}
        \begin{subfigure}[b]{.05\textwidth}
                \includegraphics[width=\linewidth]{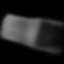}
        \end{subfigure}%
        \hspace{.001\textwidth}
        \begin{subfigure}[b]{.05\textwidth}
                \includegraphics[width=\linewidth]{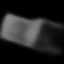}
        \end{subfigure}%
        \hspace{.001\textwidth}
        \begin{subfigure}[b]{.05\textwidth}
                \includegraphics[width=\linewidth]{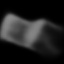}
        \end{subfigure}%
        \hspace{.001\textwidth}
        \begin{subfigure}[b]{.05\textwidth}
                \includegraphics[width=\linewidth]{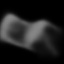}
        \end{subfigure}%
        \hspace{.001\textwidth}
        \begin{subfigure}[b]{.05\textwidth}
                \includegraphics[width=\linewidth]{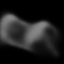}
        \end{subfigure}%
    \end{center}
    \begin{center}
        \begin{subfigure}[b]{.05\textwidth}
                \includegraphics[width=\linewidth]{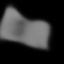}
        \end{subfigure}%
        \hspace{.001\textwidth}
        \begin{subfigure}[b]{.05\textwidth}
                \includegraphics[width=\linewidth]{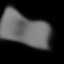}
        \end{subfigure}%
        \hspace{.001\textwidth}
        \begin{subfigure}[b]{.05\textwidth}
                \includegraphics[width=\linewidth]{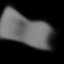}
        \end{subfigure}%
        \hspace{.001\textwidth}
        \begin{subfigure}[b]{.05\textwidth}
                \includegraphics[width=\linewidth]{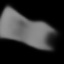}
        \end{subfigure}%
        \hspace{.001\textwidth}
        \begin{subfigure}[b]{.05\textwidth}
                \includegraphics[width=\linewidth]{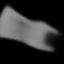}
        \end{subfigure}%
        \hspace{.001\textwidth}
        \begin{subfigure}[b]{.05\textwidth}
                \includegraphics[width=\linewidth]{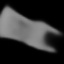}
        \end{subfigure}%
        \hspace{.001\textwidth}
        \begin{subfigure}[b]{.05\textwidth}
                \includegraphics[width=\linewidth]{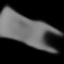}
        \end{subfigure}%
        \hspace{.001\textwidth}
        \begin{subfigure}[b]{.05\textwidth}
                \includegraphics[width=\linewidth]{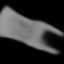}
        \end{subfigure}%
    \end{center}
    \begin{center}
        \begin{subfigure}[b]{.05\textwidth}
                \includegraphics[width=\linewidth]{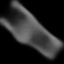}
        \end{subfigure}%
        \hspace{.001\textwidth}
        \begin{subfigure}[b]{.05\textwidth}
                \includegraphics[width=\linewidth]{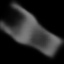}
        \end{subfigure}%
        \hspace{.001\textwidth}
        \begin{subfigure}[b]{.05\textwidth}
                \includegraphics[width=\linewidth]{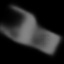}
        \end{subfigure}%
        \hspace{.001\textwidth}
        \begin{subfigure}[b]{.05\textwidth}
                \includegraphics[width=\linewidth]{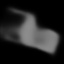}
        \end{subfigure}%
        \hspace{.001\textwidth}
        \begin{subfigure}[b]{.05\textwidth}
                \includegraphics[width=\linewidth]{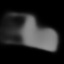}
        \end{subfigure}%
        \hspace{.001\textwidth}
        \begin{subfigure}[b]{.05\textwidth}
                \includegraphics[width=\linewidth]{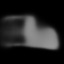}
        \end{subfigure}%
        \hspace{.001\textwidth}
        \begin{subfigure}[b]{.05\textwidth}
                \includegraphics[width=\linewidth]{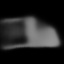}
        \end{subfigure}%
        \hspace{.001\textwidth}
        \begin{subfigure}[b]{.05\textwidth}
                \includegraphics[width=\linewidth]{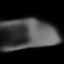}
        \end{subfigure}%
    \end{center}
    \begin{center}
        \begin{subfigure}[b]{.05\textwidth}
                \includegraphics[width=\linewidth]{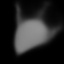}
        \end{subfigure}%
        \hspace{.001\textwidth}
        \begin{subfigure}[b]{.05\textwidth}
                \includegraphics[width=\linewidth]{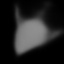}
        \end{subfigure}%
        \hspace{.001\textwidth}
        \begin{subfigure}[b]{.05\textwidth}
                \includegraphics[width=\linewidth]{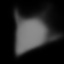}
        \end{subfigure}%
        \hspace{.001\textwidth}
        \begin{subfigure}[b]{.05\textwidth}
                \includegraphics[width=\linewidth]{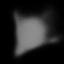}
        \end{subfigure}%
        \hspace{.001\textwidth}
        \begin{subfigure}[b]{.05\textwidth}
                \includegraphics[width=\linewidth]{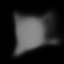}
        \end{subfigure}%
        \hspace{.001\textwidth}
        \begin{subfigure}[b]{.05\textwidth}
                \includegraphics[width=\linewidth]{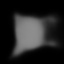}
        \end{subfigure}%
        \hspace{.001\textwidth}
        \begin{subfigure}[b]{.05\textwidth}
                \includegraphics[width=\linewidth]{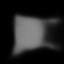}
        \end{subfigure}%
        \hspace{.001\textwidth}
        \begin{subfigure}[b]{.05\textwidth}
                \includegraphics[width=\linewidth]{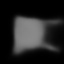}
        \end{subfigure}%
    \end{center}
    \begin{center}
        \begin{subfigure}[b]{.05\textwidth}
                \includegraphics[width=\linewidth]{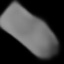}
        \end{subfigure}%
        \hspace{.001\textwidth}
        \begin{subfigure}[b]{.05\textwidth}
                \includegraphics[width=\linewidth]{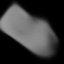}
        \end{subfigure}%
        \hspace{.001\textwidth}
        \begin{subfigure}[b]{.05\textwidth}
                \includegraphics[width=\linewidth]{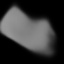}
        \end{subfigure}%
        \hspace{.001\textwidth}
        \begin{subfigure}[b]{.05\textwidth}
                \includegraphics[width=\linewidth]{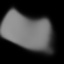}
        \end{subfigure}%
        \hspace{.001\textwidth}
        \begin{subfigure}[b]{.05\textwidth}
                \includegraphics[width=\linewidth]{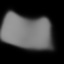}
        \end{subfigure}%
        \hspace{.001\textwidth}
        \begin{subfigure}[b]{.05\textwidth}
                \includegraphics[width=\linewidth]{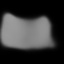}
        \end{subfigure}%
        \hspace{.001\textwidth}
        \begin{subfigure}[b]{.05\textwidth}
                \includegraphics[width=\linewidth]{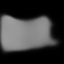}
        \end{subfigure}%
        \hspace{.001\textwidth}
        \begin{subfigure}[b]{.05\textwidth}
                \includegraphics[width=\linewidth]{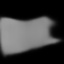}
        \end{subfigure}%
    \end{center}
    \caption{Linear interpolation between two points in temporal factor with static factor held fixed for the Rotating Chairs dataset.}
    \label{fig:temporal_interp_chair}
    \end{figure}

    In figures \ref{fig:temporal_interp} and \ref{fig:temporal_interp_chair}, we present outputs from the decoder of inputs that similarly interpolate between two sampled points in the temporal factor while holding a sample from the static factor fixed. In the Bouncing MNIST example, the appearance of the character remains constant but the character's location smoothly translates. This demonstrates that the temporal factor of the embedding encodes no information regarding the identity of the character. In the Rotating Chairs example, the chair appears to smoothly rotate while the shape of the chair remains roughly constant demonstrating that the temporal factor encodes little information regarding the shape of the chair and encodes mainly orientation information.
    
\begin{figure}
    \begin{center}
        \begin{subfigure}[b]{.225\textwidth}
                \includegraphics[width=\linewidth]{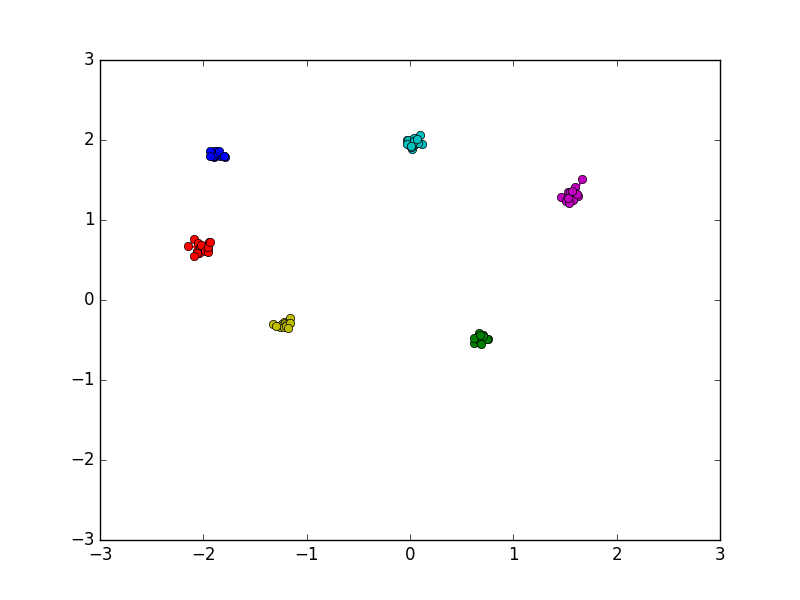}
        \end{subfigure}%
        \hspace{.001\textwidth}
        \begin{subfigure}[b]{.225\textwidth}
                \includegraphics[width=\linewidth]{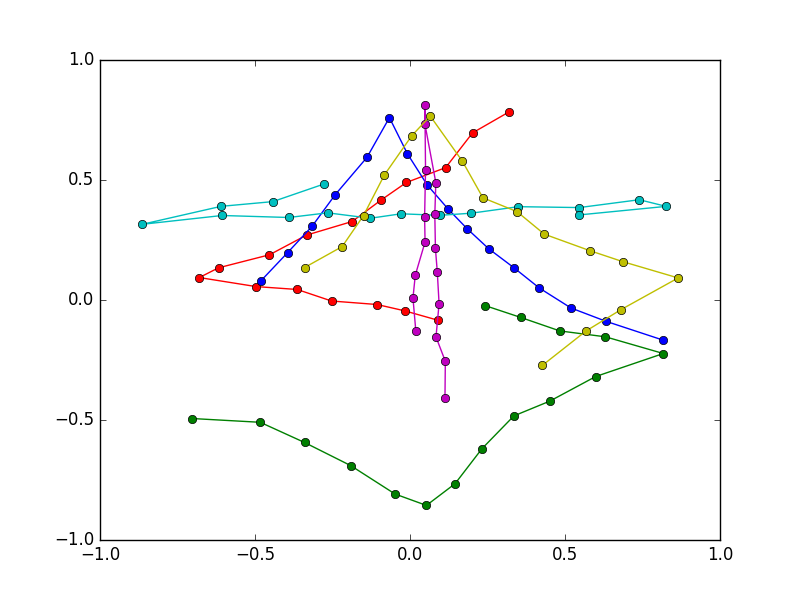}
        \end{subfigure}%

        \begin{subfigure}[b]{.225\textwidth}
                \includegraphics[width=\linewidth]{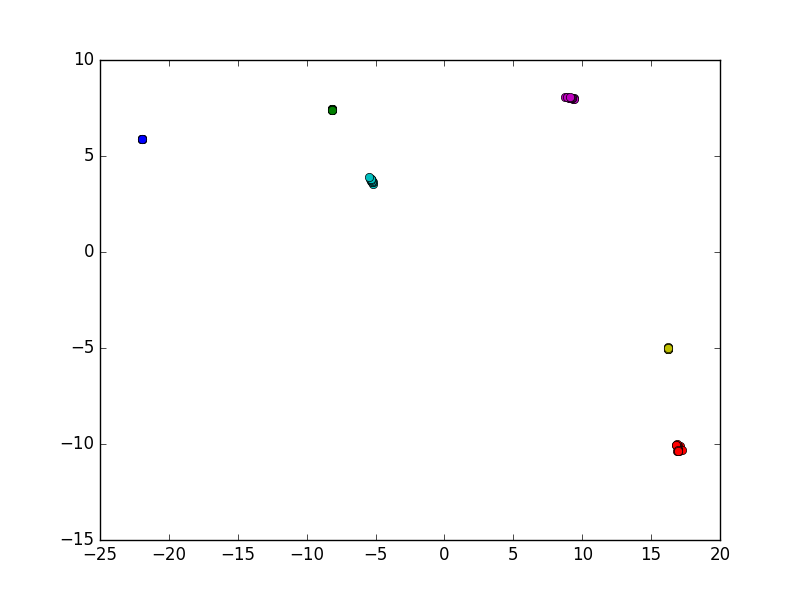}
        \end{subfigure}%
        \hspace{.001\textwidth}
        \begin{subfigure}[b]{.225\textwidth}
                \includegraphics[width=\linewidth]{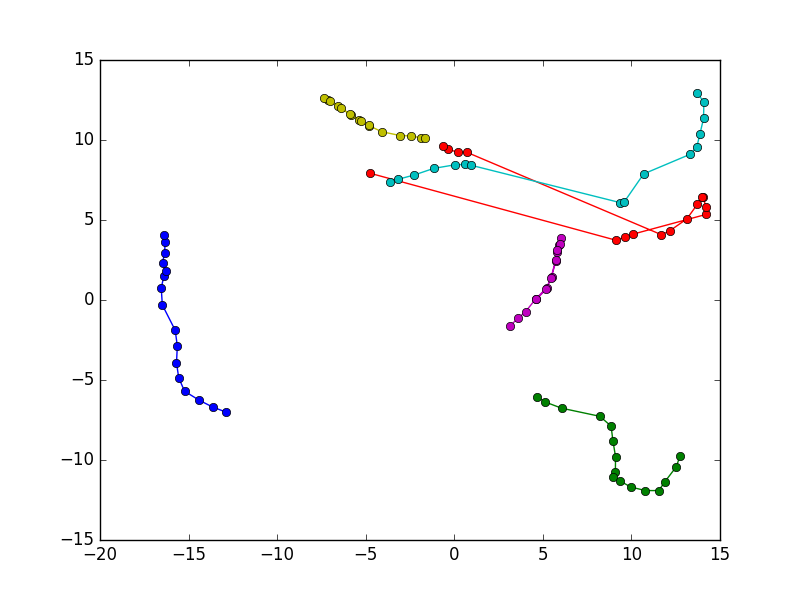}
        \end{subfigure}%
        \caption{Plots of encoded videos drawn from the testing set. The first column plots the static factor of the learned embedding and the second column plots the temporal factor of the learned embedding. Points are color-coded by which video they are drawn from. In each row, the colors are consistent across the static and temporal factors. Row 1 is our model trained with 2 features for each factor. Row 2 is our model trained with 4 features for each factor, mapped to 2-dimensions via T-SNE.}
        \label{fig:encoding_plots}
    \end{center}
    \end{figure}

    In figure \ref{fig:encoding_plots} we draw 6 videos from our testing set and sample encodings from our model's parametrized output distribution. We plot each factor separately to demonstrate the differences in their underlying distributions. Points of the same color were drawn from the same video and are connected to display the temporal order of the video frames. 
    
\subsection{Quantitative Results}
    We compare our model with the benchmark models on our disentanglement score defined in section 4.1. Table \ref{tab:quant_res} displays these results. All versions of our model outperform all benchmarks. It can be seen that as the regularization strength $\beta$ increases, all models see an increase in their disentanglement score as \cite{higgins2016early} would suggest.
    
    We run a number of transfer learning experiments to demonstrate our model's ability to learn superior features for frame classification. These results can be viewed in tables \ref{tab:stat_clf} and \ref{tab:stat_clf_chair}. Our comparison is carried out as follows. Given a dataset $D$, a number of features $n$, and a regularization strength $k$. We then fit a VAE and our slow feature benchmark models with $n$ features and regularization strength $k$ on $D$. We fit our model where each factor has $n$ features and regularization strength $k$ on $D$. 
    
    For the benchmarks, we encode $D$ with our trained models to obtain features $H$ and fit an $l2$-svm on $H$ to targets $y$ and present the results. 
    
    For our model, we encode $D$ to obtain features $H_s$ and $H_t$ (for each factor) then train an $l2$-svm with targets $y$ on $H_s$ and present the results. It should be noted that by using only half of the features, our model has exactly the same number of parameters as the benchmarks used for comparison. The success of our model indicates that our prior allows each factor to better focus on the static and temporal information. 
    
    In these experiments, we use our static classification targets for each dataset as they require more high level features than our temporal classification targets to accurately classify. 
    
    In both the Bouncing MNIST (both encoding sizes) and the Rotating Chair experiments, our model performs the best. We believe this is due to the fact that our prior allows the model to dedicate the features used for transfer only to the static factor where other models need to share their capacity with the temporal information that our model ignores. 
    
    We found that in our models with more latent features, the results were very sensitive to the regularization strength parameter $\beta$. At the bayes solution of $\beta = 1$ we observed little or no disentanglement. In these models, the temporal factor was mainly used to encode the data indicating that static features are much more challenging to learn than slow features. This is best illustrated by the results of table \ref{tab:stat_clf_chair}. We see that at the bayes solution, our model greatly under-performs the benchmarks, but with proper variational regularization, our model's performance begins to surpass the benchmarks. We believe that the performance would continue to increase with $\beta$. 
     
    This indicates that learning factored representations may be a much harder problem than simply encoding data.

    
    \begin{table}
    \begin{center}
    \begin{tabular}{|c|c|c|c|}
    \hline
    Model & KL-weight & D-Score MNIST & D-Score Chairs\\
    \hline\hline
    Ours & 1 & 6.71 & 1.79\\
    \hline
    Ours & 2 & 6.74 & 1.73\\
    \hline
    Ours & 4 & $\mathbf{6.82}$ & $\mathbf{1.81}$ \\
    \hline\hline
    Slow & 1 & 4.70 & 1.19\\
    \hline
    Slow & 2 & 5.78 & 1.57\\
    \hline
    Slow & 4 & 6.38 & 1.39\\
    \hline\hline
    VAE & 1 & 1.08 & 1.27\\
    \hline
    VAE & 2 & 1.70 & 1.24\\
    \hline
    VAE & 4 & 1.71 & 1.35\\
    \hline
    \end{tabular}
    \end{center}
    \caption{Disentanglement Results. Disentanglement score is the metric defined in section 4.1. All MNIST models have 4 latent features (total) and all Rotating Chair models have 8 latent features. Results for our model are taken over the natural static-temporal factoring and results for other models are taken as the maximum over all subsets of features.}
    \label{tab:quant_res}
    \end{table}
    
    \begin{table}
    \begin{center}
    \begin{tabular}{|c|c|c|c|}
    \hline
    Model & KL-weight & Acc 4 & Acc 2 \\
    \hline\hline
    Ours & 1 & 0.83 & $\mathbf{0.71}$ \\
    \hline
    Ours & 2 & $\mathbf{0.88}$ & 0.64 \\
    \hline
    Ours & 4 & 0.82 & 0.69 \\
    \hline\hline
    Slow & 1 & 0.58 & 0.11 \\
    \hline
    Slow & 2 & 0.65 & 0.12 \\
    \hline
    Slow & 4 & 0.66 & 0.13 \\
    \hline\hline
    VAE & 1 & 0.80 & 0.13 \\
    \hline
    VAE & 2 & 0.52 & 0.12 \\
    \hline
    VAE & 4 & 0.51 & 0.13 \\
    \hline

    \end{tabular}
    \end{center}
    \caption{Bouncing MNIST static classification results.``Acc $x$'' indicates the classification accuracy of a linear classifier trained on a feature subset of size $x$.}
    \label{tab:stat_clf}
    \end{table}
    
    \begin{table}
    \begin{center}
    \begin{tabular}{|c|c|c|}
    \hline
    Model & KL-weight & Accuracy \\
    \hline\hline
    Ours & 1 & 0.33 \\
    \hline
    Ours & 2 & 0.48 \\
    \hline
    Ours & 4 & $\mathbf{0.59}$ \\
    \hline\hline
    Slow & 1 & 0.24 \\
    \hline
    Slow & 2 & 0.34 \\
    \hline
    Slow & 4 & 0.37 \\
    \hline\hline
    VAE & 1 & 0.52 \\
    \hline
    VAE & 2 & 0.51 \\
    \hline
    VAE & 4 & 0.54 \\
    \hline

    \end{tabular}
    \end{center}
    \caption{Rotating Chairs static classification results. Our model trained with 4 features in each factor and benchmarks trained with 4 features.}
    \label{tab:stat_clf_chair}
    \end{table}
    


\section{Conclusions and Further Work}
    In this paper we  have presented a neural network model that learns to decompose the static and temporally varying semantic information in video. We have demonstrated the success of this model in factoring its representation both quantitatively and qualitatively.
    
    We are aware that our model is sensitive to some of its hyper-parameters; mainly $\beta$ and the size of the latent representation. We are interested in carrying out more experiments to better understand their relationship and their effect on the semantics of the model's learned representation. 
    
    We feel that there is much room for further work extending this model. We are interested in extensions which operate on larger-scale and more natural video datasets. We are aware that it is a rather naive assumption that information in a video perfectly factors into static and temporal information. This work could be extended to use a prior with multiple factors such that each factor changes at varying rates from fast to slow to static. We believe this will more accurately describe the information flow in real-world videos.
    
    We are also interested in exploring other kinds of assumptions, such as sparsity and orthogonality, that can be encoded into priors to be enforced upon the model. There is a wealth of previous work on such properties that have yet to be applied to factored representation learning.

\section{Acknowledgements}
    This work was performed under the auspices of the U.S. Department of Energy by Lawrence Livermore National Laboratory under Contract DE-AC52-07NA27344 (LLNL-CONF-677443). Funding partially provided by LDRD 17-SI-003.

    \begin{table}
    \begin{center}
    \begin{tabular}{|c|c|c|c|c|c|}
    \hline
    Layer & Type & BN & \# Out & Size & Stride  \\
    \hline\hline
    Conv 1a & conv & \checkmark & 16 & 3 & 1 \\
    Conv 1b & conv & \checkmark & 16 & 3 & 2 \\
    Conv 2a & conv & \checkmark & 32 & 3 & 1 \\
    Conv 2b & conv & \checkmark & 32 & 3 & 2 \\
    Conv 3a & conv & \checkmark & 64 & 3 & 1 \\
    Conv 3b & conv & \checkmark & 64 & 3 & 2 \\
    Conv 4a & conv & \checkmark & 128 & 3 & 1 \\
    Conv 4b & conv & \checkmark & 128 & 3 & 2 \\
    Linear 5 & linear & \checkmark & 256 & & \\
    out $\mu$      & linear &  & (4,8) & & \\
    out $\sigma$   & linear &  & (4,8) & & \\
    \hline
    \end{tabular}
    \end{center}
    \caption{Encoder Model Architecture. ``BN'' indicates batch normalization.}
    \label{tab:enc_arch}
    \end{table}
    
    
    \begin{table}
    \begin{center}
    \begin{tabular}{|c|c|c|c|c|c|}
    \hline
    Layer & Type & BN & \# Out & Size & Stride  \\
    \hline\hline
    Linear 1 & linear & \checkmark & 256 & & \\
    Linear 2 & linear & \checkmark & 2048 & & \\
    $\mbox{Conv}^T$ 1a & $\mbox{conv}^T$ & \checkmark & 128 & 3 & 1 \\
    $\mbox{Conv}^T$ 1b & $\mbox{conv}^T$ & \checkmark & 64 & 3 & 2 \\
    $\mbox{Conv}^T$ 2a & $\mbox{conv}^T$ & \checkmark & 64 & 3 & 1 \\
    $\mbox{Conv}^T$ 2b & $\mbox{conv}^T$ & \checkmark & 32 & 3 & 2 \\
    $\mbox{Conv}^T$ 3a & $\mbox{conv}^T$ & \checkmark & 32 & 3 & 1 \\
    $\mbox{Conv}^T$ 3b & $\mbox{conv}^T$ & \checkmark & 16 & 3 & 2 \\
    $\mbox{Conv}^T$ 4a & $\mbox{conv}^T$ & \checkmark & 16 & 3 & 1 \\
    $\mbox{Conv}^T$ 4b & $\mbox{conv}^T$ &  & 1 & 3 & 2 \\
    \hline
    \end{tabular}
    \end{center}
    \caption{Decoder Model Architecture. ``$\text{conv}^T$'' indicates a transpose convolutional layer. ``BN'' indicates batch normalization.}
    \label{tab:dec_arch}
    \end{table}


\bibliographystyle{ieee}
\bibliography{egbib}

\end{document}